\documentclass[a4paper, 10pt, conference]{cssconf}        
\pdfoutput=1
\IEEEoverridecommandlockouts                              
\usepackage{graphicx} 
\usepackage{times} 
\usepackage{amsmath} 
\usepackage{amssymb}  
\usepackage{bm}  

\usepackage{multirow,color}
\usepackage{hyperref}
\DeclareMathAlphabet{\bdmath}{OML}{cmm}{b}{it}     
\mathversion{normal}

\def\0{\mathbf{0}}

\def\F{\bdmath{F}}
\def\J{\bdmath{J}}
\def\K{\bdmath{K}}

\def\W{\bdmath{W}}

\def\f{\bdmath{f}}

\def\q{\bdmath{q}}
\def\u{\bdmath{u}}
\def\x{\bdmath{x}}

\def\qdot{\dot{\q}}

\sloppypar

\title{\LARGE \bf
Non-linear Task-Space Disturbance Observer for Position Regulation of Redundant Robot Arms against Perturbations in 3D Environments}

\author{Tapomayukh Bhattacharjee, Yonghwan Oh* and Sang-Rok Oh
\thanks{During this work, T. Bhattacharjee was with the Interaction and Robotics Research Center, KIST, 39-1 Hawolgok-dong, Wolsong Gil 5, Seongbuk-gu, Seoul 136-791, South-Korea. Now, he is with Georgia Institute of Technology, USA}
\thanks{Y. Oh, and S. R. Oh are with the Interaction and Robotics Research Center, KIST, 39-1 Hawolgok-dong, Wolsong Gil 5,
    Seongbuk-gu, Seoul 136-791, South-Korea
        {\tt\small \{oyh, sroh\}@kist.re.kr}}%
\thanks{*Y. Oh is the corresponding author}
}

\begin{document}

\maketitle
\thispagestyle{empty}
\pagestyle{empty}

\begin{abstract}
Many day-to-day activities require the dexterous manipulation of a redundant humanoid arm in complex 3D environments. However, position regulation of such robot arm systems becomes very difficult in presence of non-linear uncertainties in the system. Also, perturbations exist due to various unwanted interactions with obstacles for clumsy environments in which obstacle avoidance is not possible, and this makes position regulation even more difficult. This report proposes a non-linear task-space disturbance observer by virtue of which position regulation of such robotic systems can be achieved in spite of such perturbations and uncertainties. Simulations are conducted using a 7-DOF redundant robot arm system to show the effectiveness of the proposed method. These results are then compared with the case of a conventional mass-damper based task-space disturbance observer to show the enhancement in performance using the developed concept. This proposed method is then applied to a controller which exhibits human-like motion characteristics for reaching a target. Arbitrary perturbations in the form of interactions with obstacles are introduced in its path. Results show that the robot end-effector successfully continues to move in its path of a human-like quasi-straight trajectory even if the joint trajectories deviated by a considerable amount due to the perturbations. These results are also compared with that of the unperturbed motion of the robot which further prove the significance of the developed scheme.
\end{abstract}

\newcommand\pkg[1]{\textsf{#1}}
\newcommand\file[1]{\texttt{#1}}

\newcommand{\norm}[1]{\left\lVert#1\right\rVert}


\section{INTRODUCTION}

Humanoid robots have increasingly been used in complex environments to make the use of such robots more feasible in human life. One of the most important aspects in these environments is, therefore, the dexterous manipulation of humanoid redundant arm systems. Redundancy is almost a prerequisite in such cases because it can be used to manipulate objects in human environments dexterously. However, the motion of any such redundant arm systems is complex mainly because there is no one-to-one correspondence between the task-space and joint-space of the robot. This is because when the task-space motions are mapped to joint-space to input the required commands to the robots, the mapping is not unique and hence, many solutions exist. This, however, poses an opportunity to select a single solution out of those multiple solutions which satisfies some of the other possible task requirements such as obstacle avoidance and specific motion characteristics. Thus, several studies have been conducted to utilize this aspect of redundancy in the manipulation of objects in complex environments\,\cite{1}-\cite{5}.

Many studies\,\cite{6}-\cite{7-2} attempt to control the motion of redundant systems by segregating the motion space into two subspaces such as the task-space motion component and the null-space motion component and by controlling these according to desired performance objectives. One of the advantages of this method is that the task-space component can be commanded to reach a desired task objective while the null-space component can be separately controlled for redundancy resolution in order to enhance a specific performance criteria without affecting the task-space objective. However, in spite of this promising method, position regulation in complex dynamic environments full of obstacles has not yet been completely successful mainly because of the highly non-linear nature of uncertainties present in the robot dynamics as well as the inability to avoid all the obstacles in such proximity. Though model-based uncertainty compensation techniques and model-less disturbance observers have been used by researchers\,\cite{8}-\cite{12}, these methods have still not been able to achieve the desired performance largely because of inaccurate models or non-linear nature of robot dynamics.

This report hence employs a non-linear disturbance observer in the task-space for position regulation of redundant robot arms in spite of interactions with obstacles in complex 3D environments wherein obstacle avoidance is not possible. We, however, assume that no obstacles lie directly on the trajectory of the end-effector in which case obstacle avoidance is a must. However, obstacles may lie very near to the trajectory and hence, the robot's non-terminal links might interact with the obstacles depending on the configuration of the robot. We also assume that the interactions with the obstacles do not affect the stability of the robot and only changes the joint trajectory. This assumption is quite practical for soft environments and compliant robot arms. Henceforth, we shall refer to these unwanted interactions as perturbations in the environment. We have used a 7-DOF redundant robot as our simulation test bed. A modified version of the Newton-Euler Algorithm has been implemented to compute the detailed non-linearities of the robot dynamics and to use it in the task-space disturbance observer. Results verify the efficacy of our algorithm for effective position regulation in such complex environments. In addition to this, this report also compares these results with that of a conventional task-space disturbance observer with a simplistic mass-damper lumped model as the nominal model without considering the non-linearities. Simulations are then conducted in a real-time simulation software called {\it RoboticsLab} and the results show the efficacy of the proposed scheme as compared to the conventional scheme. \textit{The links to the video of the above simulation result is provided in the Section \ref{sec:result}.}

To further prove the significance of the proposed method, it is applied to a controller which exhibits human-like characteristics for reaching motion. Arbitrary perturbations are introduced in the robotic system which affect the joint trajectories of the robot arm. However, results show that the robot end-effector successfully continues to move along the human-like quasi-straight trajectory for reaching its target using the proposed scheme in spite of the perturbations. These results are also compared to the unperturbed motion of the robot arm which show the significance of the developed method.

\section{POSITION REGULATION USING TASK-SPACE DYNAMICS}

Position regulation, generally, becomes difficult if perturbations are present in an environment. However, it is possible to maintain a particular position in spite of these perturbations in the joint space by utilizing the dynamics of the inherent redundancy in a system. A perturbation in an environment can be usually represented as an external force/torque acting on the system. Hence, theoretically if any external force applied changes only the configuration of the internal robot joints without moving the end-effector, accurate position regulation might be possible provided all other non-linear uncertainties in the robotic system are well compensated. The next sub-sections on robot dynamics will show us as to how this can be achieved.

\subsubsection{Joint-Space Dynamics}

Any robot manipulator operating in real environment is affected by the inertial, coriolis, gravity, and friction effects as well as external force influence. Therefore, the general dynamic equation of any robot behavior can be written in joint-space in the form of

\begin{equation}
\bm{\tau}  = \bm{M}(\bm{q})\bm{\ddot q} + \bm{c}(\bm{q},\bm{\dot q}) + \bm{g}(\bm{q}) + \bm{f}(\bm{q}) + \bm{{\tau}_e}
 \label{eq:1}
\end{equation}

where $\bm{M}(\bm{q})$ denotes a symmetric, positive definite robot inertia matrix, $\bm{c}(\bm{q},\bm{\dot q})$ is the Coriolis and Centrifugal Vector, $\bm{g}(\bm{q})$ is the gravity vector, $\bm{f}(\bm{q})$ is the friction component and $\bm{{\tau}_e}$ is the external force acting on the manipulator joint. The Coriolis and Centrifugal term in (\ref{eq:1}) can be factorized as

\begin{equation}
\bm{c}(\bm{q},\bm{\dot q}) = \bm{C}(\bm{q},\bm{\dot q})\bm{\dot q} \label{eq:2}
\end{equation}

where $\bm{C}(\bm{q},\bm{\dot q})$ holds the skew-symmetric property given in (\ref{eq:3}) and is calculated from Christoffel symbols of the second kind.

\begin{equation}
{\bm{x}^T}\left( {\bm{\dot M}(\bm{q}) - 2\bm{C}(\bm{q},\bm{\dot q})} \right)\bm{x} = \bm{0},\quad \forall \bm{x} \in {\mathbb{R}^N} \label{eq:3}
\end{equation}

Hence, the resultant general dynamic equation is given by

\begin{equation}
\bm{\tau}  = \bm{M}(\bm{q})\bm{\ddot q} + \bm{C}(\bm{q},\bm{\dot q})\bm{\dot q} + \bm{g}(\bm{q}) + \bm{f}(\bm{q}) + \bm{{\tau}_e} \label{eq:4}
\end{equation}

If we choose to model the redundant robot manipulator as a simple linear mass-damper system such as $\bm{M_s}\bm{\ddot q} + \bm{B_s}\bm{\dot q}$, then the disturbance torque $\bm{{\tau}_d}$ to the system is given by (\ref{eq:5}) which will be eventually canceled by any joint-space disturbance observer, if applied.

\begin{equation}
\bm{{\tau}_d}  = (\bm{M}(\bm{q})-\bm{M_s})\bm{\ddot q} + (\bm{C}(\bm{q},\bm{\dot q})-\bm{B_s})\bm{\dot q} + \bm{g}(\bm{q}) + \bm{f}(\bm{q}) + \bm{{\tau}_e} \label{eq:5}
\end{equation}

Hence, as evident from (\ref{eq:5}), any disturbance observer in joint-space cancels the external force/torque because it is considered as a disturbance itself along with the variation in inertial parameters, Coriolis, friction and gravity effects. If this were to be implemented in perturbed environments, theoretically, the robot arm will continue to hit the obstacles subsequently after it hits those obstacles for the first time because it will tend to come back to its original configuration just before it was perturbed. Therefore, it is necessary to come up with an alternative formulation through which position regulation is possible and the task-space objectives can be achieved in spite of the presence of unwanted interactions with obstacles and non-linear uncertainties in the system.

\subsubsection{Task-Space Dynamics}

An alternative can be to express the dynamics using the task-space formulation and apply a disturbance observer in the operational space which can cancel the disturbances in the task-space only. The general dynamic equation of a multi degree-of-freedom robot manipulator in the task-space can be derived from the joint-space equation shown in (\ref{eq:5}) as given in (\ref{eq:6}) and (\ref{eq:7})\,\cite{7,8}. Please note that the friction effects are neglected.

\begin{equation}
\bm{f} = \bm{\Lambda}(\bm{q})\bm{\ddot x} + \bm{\Gamma}(\bm{q},\bm{\dot q})\bm{\dot x} + \bm{\eta}(\bm{q}) + \bm{f_e} \label{eq:6}
\end{equation}

where

\begin{equation}
\begin{gathered}
  \bm{\Lambda}(\bm{q}) = {\bm{J}_M^{+T}}\bm{M}{\bm{J}_M^{+}}, \hfill \\
  \bm{\Gamma}(\bm{q},\bm{\dot q}) = {\bm{J}_M^{+T}}\left[ {\bm{C}(\bm{q},\bm{\dot q}) - \bm{M}{\bm{J}_M^{+}}\bm{\dot J}} \right]{\bm{J}_M^{+}}, \hfill \\
  \bm{\eta}(\bm{q}) = {\bm{J}_M^{+T}}\bm{g}(\bm{q}). \hfill \\
\end{gathered} \label{eq:7}
\end{equation}

and $\bm{f}$ represents the contribution of the end-effector forces due to joint-actuation and $\bm{f_e}$ represents the external forces or the contribution of the end-effector forces due to contact with the environment. $\bm{J}$ is the Jacobian matrix and $\bm{J}_M^{+}$ is the Moore-Penrose pseudoinverse of the Jacobian matrix which is weighted by the inverse of the inertia matrix $\bm{M}^{ -1}$. By substituting (\ref{eq:7}) in (\ref{eq:6}) and by expressing $\bm{\ddot x} = \bm{J}\bm{\ddot q} + \bm{\dot J}\bm{\dot q}$ and $\bm{\dot q} = {\bm{J}_M^{+}}\bm{\dot x}$, we have the following.

\begin{equation}
\bm{f} = \bm{\Lambda}(\bm{q})\bm{J}\bm{\ddot q} + {\bm{J}_M^{+T}}\bm{C}(\bm{q},\bm{\dot q})\bm{\dot q} + {\bm{J}_M^{+T}}\bm{g} + \bm{f_e}
\label{eq:8}
\end{equation}

which can be rewritten from the relation $\bm{\Lambda}(\bm{q}) = {\bm{J}_M^{+T}}\bm{M}{\bm{J}_M^{+}}$ as given below.

\begin{equation}
\bm{f} = {\bm{J}_M^{+T}}\left( {\bm{M}(\bm{q})\bm{\ddot q} + \bm{C}(\bm{q},\bm{\dot q})\bm{\dot q} + \bm{g}(\bm{q})} \right) + \bm{{f_e}}
\label{eq:9}
\end{equation}

Therefore, if we neglect the friction forces, then from (\ref{eq:4}) and (\ref{eq:9}), we have (\ref{eq:10}).

\begin{equation}
\bm{f} = \bm{J}_M^{+T}\left( \bm{\tau}  - \bm{{\tau}_e} \right) + \bm{f_e}\; \label{eq:10}
\end{equation}

Since, ${\bm{J}_M^{+T}}\bm{{\tau} _e} = \bm{{f_e}}$, hence,

\begin{equation}
\bm{f} = {\bm{J}_M^{+T}}\bm{\tau} \label{eq:11}
\end{equation}

When the above task-space command input is mapped back to joint-space, a null-space projection component arises in redundant manipulators and hence, the general solution of (\ref{eq:11}) is given by (\ref{eq:12}). This is because ${\bm{J}^{T}}$ is the right pseudo-inverse of ${\bm{J}_M^{+T}}$ weighted by ${\bm{M}^{ - 1}}$ \,\cite{9}. The second term in (\ref{eq:12}) represents the required null-space projection component.

\begin{equation}
\bm{\tau}  = {\bm{J}^T}\bm{f} + \left( {\bm{I} - {\bm{J}^T}(\bm{q}){\bm{J}_M^{+T}}(\bm{q})} \right)\;\bm{{\tau} _0}
 \label{eq:12}
\end{equation}

where $\bm{{\tau}_0}$ is a vector of arbitrary torques which does not contribute to end-effector forces and motion. Now, if a disturbance observer is applied in the task-space, (\ref{eq:12}) is modified to (\ref{eq:13}).

\begin{equation}
\bm{\tau}  = {\bm{J}^T}(\bm{f}-\bm{f_d}) + \left( {\bm{I} - {\bm{J}^T}(\bm{q}){\bm{J}_M^{+T}}(\bm{q})} \right)\;\bm{{\tau}_0}
 \label{eq:13}
\end{equation}

where $\bm{f_d}$ is given by (\ref{eq:14}) if the nominal model of the robot manipulator is given by $\bm{M_s^{'}}\bm{\ddot x} + \bm{B_s^{'}}\bm{\dot x}$.

\begin{equation}
\bm{f_d} = \left( {\bm{\Lambda}(\bm{q}) - \bm{M_s^{'}}} \right)\bm{\ddot x} + \left( {\bm{\Gamma}(\bm{q},\bm{\dot q}) - \bm{B_s^{'}}} \right)\bm{\dot x} + \bm{\eta}(\bm{q}) + \bm{f_e} \label{eq:14}
\end{equation}

Therefore, (\ref{eq:13}) becomes (\ref{eq:15}) or (\ref{eq:16}) and thus the external torques acting on the robot-manipulator due to unwanted interactions with obstacles, can be isolated. These can, eventually, be cancelled by applying a non-linear task-space disturbance observer and hence, position regulation can be achieved in spite of such unwanted perturbations.

\begin{equation}
\left( {\bm{I} - {\bm{J}^T}(\bm{q}){\bm{J}_M^{+T}}(\bm{q})} \right)\bm{{\tau} _0} = \bm{\tau}  - {\bm{J}^T}\left( {\bm{M_s^{'}}\bm{\ddot x} + \bm{B_s^{'}}\bm{\dot x}} \right) \label{eq:15}
\end{equation}

or

\begin{equation}
\bm{{\tau}_0} = {\left( {\bm{I} - {\bm{J}^T}(\bm{q}){\bm{J}_M^{+T}}(\bm{q})} \right)_M^{+}}\left[ {\bm{\tau}  - {\bm{J}^T}\left( {\bm{M_s^{'}}\bm{\ddot x} + \bm{B_s^{'}}\bm{\dot x}} \right)} \right] \label{eq:16}
\end{equation}

Please note that the conversion from (\ref{eq:15}) to (\ref{eq:16}) might not be possible using standard weighted pseudo-inverse techniques as the rank of the matrix is equal to the degree of redundancy. However, it can be achieved using Singular Value Decomposition techniques.

\section{NON-LINEAR TASK-SPACE DISTURBANCE OBSERVER}

\begin{figure}
    \centering%
    \includegraphics[width=0.8\columnwidth,keepaspectratio,clip]{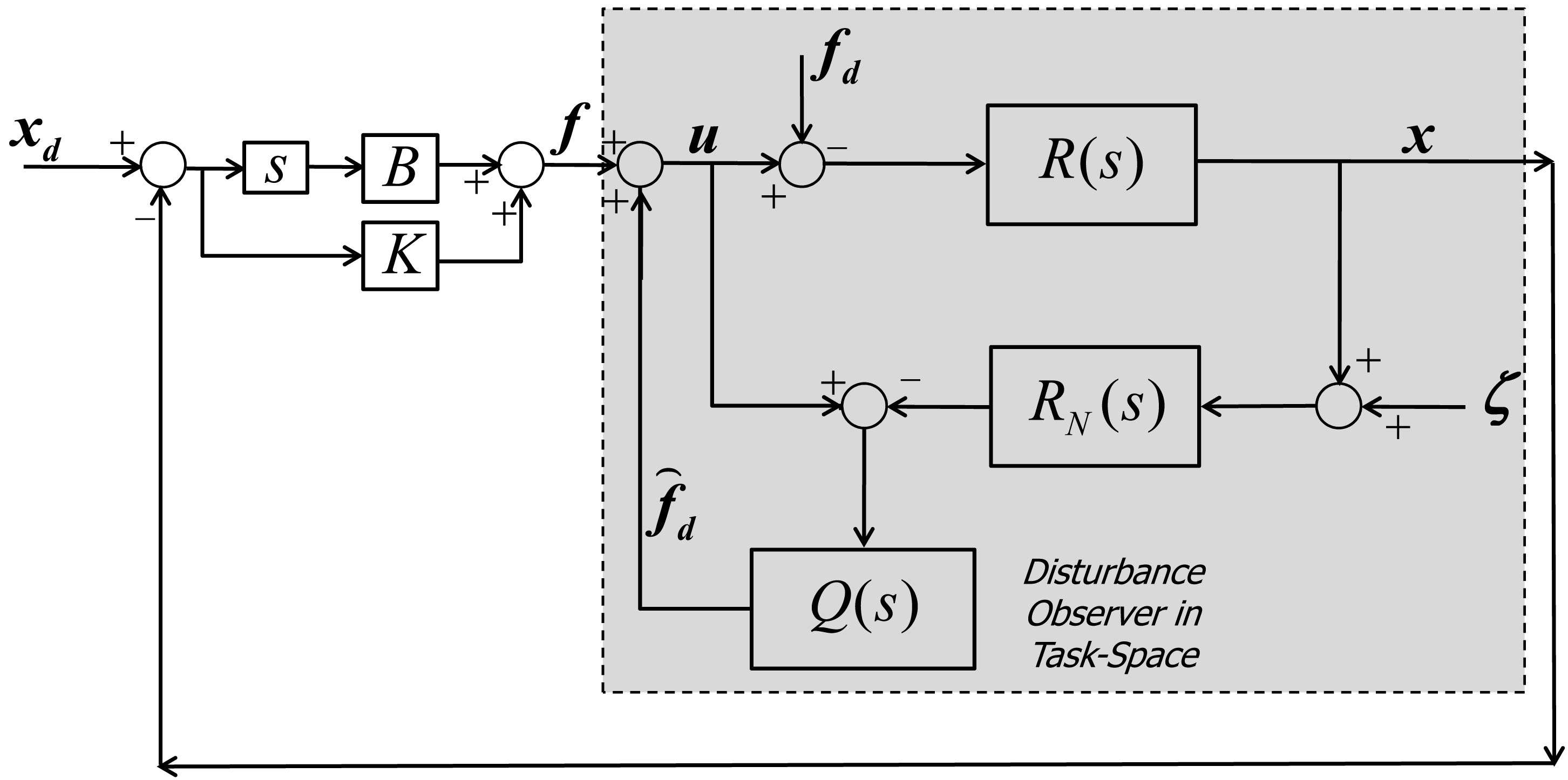}
    \caption{Model of a non-linear task-space disturbance observer with PD-control input.}        \label{fig:0}
\end{figure}

A disturbance observer applied in the operational workspace or task-space is called a task-space disturbance observer\,\cite{10}. The model of a non-linear task-space disturbance observer is shown as the shaded region in Fig.\,\ref{fig:0} where $\bm{f}$ is the reference force input, $\bm{x}$ is the output, $\bm{\zeta}$ is the noise input, and $\bm{f_d}$ is the disturbance force. $R(s)$ is the actual model of the robot manipulator while $R_N(s)$ is the non-linear nominal model. $Q(s)$ is the low-pass Q-filter of the disturbance observer. $\bm{u}$ is the command input while $\bm{{\overset{\lower0.5em\hbox{$\smash{\scriptscriptstyle\frown}$}}{{f}} _d}}$ is the estimated disturbance from the disturbance observer. The non-linear disturbance observer estimates the disturbance force and nullifies its effect by adding it to the reference force input.

The design of the low-pass filter is the most crucial element in the disturbance observer design. The sensitivity functions of the above disturbance observer are given by (\ref{eq:17}) which show that if the nominal model is very accurate such as $R = R_N$, which is only possible if all the detailed non-linearities are modeled, then the low-pass filter is the only deciding factor of the performance of the disturbance observer.

\begin{equation}
\begin{gathered}
  T = \frac{{QR}}
{{Q(R - {R_N}) + {R_N}}} \hfill \\
  S = \frac{{{R_N}(1 - Q)}}
{{Q(R - {R_N}) + {R_N}}} \hfill \\
\end{gathered} \label{eq:17}
\end{equation}

Therefore, the selection of the order of the low-pass filter is very important. An optimal selection of the order is very important because, if the number of integrators $(1/s)$ is high, the disturbance rejection property gets enhanced for low frequencies while it results in more peaking of the sensitivity functions\,\cite{11}. Hence, we have optimally selected a 3rd-order Q-filter as given below considering all the above aspects \,\cite{12}.

\begin{equation}
Q(s) = \frac{{3{\tau _s} + 1}}
{{{{({\tau _s})}^3} + 3{{({\tau _s})}^2} + 3({\tau _s}) + 1}} \label{eq:18}
\end{equation}

The reference force input $\bm{f}$ and the disturbance force $\bm{f_d}$ are given by (\ref{eq:6}) and (\ref{eq:14}) respectively. The estimated disturbance force $\bm{{\overset{\lower0.5em\hbox{$\smash{\scriptscriptstyle\frown}$}}{f} _d}}$, is computed by the non-linear disturbance observer by using the information of the input force as well as the output and is given in (\ref{eq:19}).

\begin{equation}
\bm{{\overset{\lower0.5em\hbox{$\smash{\scriptscriptstyle\frown}$}}{f}_d}} = \bm{f}\left( {\frac{{Q(s)}}
{{1 - Q(s)}}} \right) - \bm{x}\left( {\frac{{R_N^{ - 1}(s)Q(s)}}
{{1 - Q(s)}}} \right) \label{eq:19}
\end{equation}

The above estimated force is then added to cancel the actual disturbance force effects in the task-space for the redundant robot manipulator. Also, it is to be noted that the above disturbance observer is applied in a decentralized way for each of the three directions of the motion such as $X$, $Y$, and $Z$ axes.

\section{SYSTEM MODELING}

\begin{figure}
    \centering%
    \includegraphics[width=0.3\columnwidth,keepaspectratio,clip]{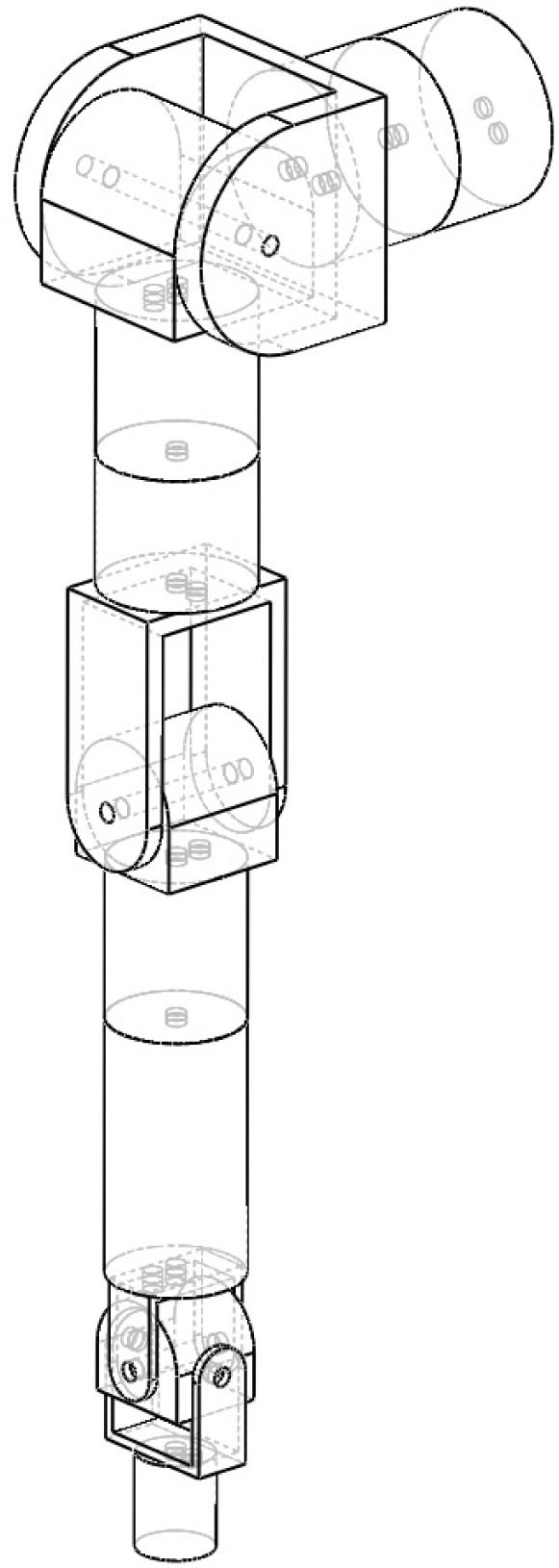}
    \caption{A 7-DOF redundant humanoid robot arm model.}        \label{fig:2}
\end{figure}

In this report, we have considered a 7-DOF redundant robot arm for analysis purposes as shown in Fig.\,\ref{fig:2}.
Let~$\bm{I_{i}}$ denote the inertia tensors for all the links of the robot wherein $i = 1, \ldots, 7$ and the diagonal elements are given by $I_{xx},~I_{yy}$, and $I_{zz}$ which are the principal moments of inertia. The non-diagonal elements given by $I_{xy},~I_{yz}$, and $I_{xz}$ are negligible for highly symmetric construction of links. Let $\ell_i$ and $m_i$ denote the link lengths and link masses respectively. The lengths of the links as well as the dynamic parameters such as the link inertia and mass properties are based on rough human measurements, and are given in Table~\ref{tab:1}. All dimensions used in this report are in S.I units.
\begin{table}
\caption{Robot parameters}  \label{tab:1}
\begin{center}
\begin{tabular}{|c|c|c|c|c|c|c|c|}    \hline
        & \multirow{2}{*}{}Joint & Joint & Joint & Joint & Joint & Joint & Joint\\
        & 1 & 2 & 3 & 4 & 5 & 6 & 7                           \\\hline\hline
    $I_{xx}$    & 0.004 & 0.012 & 0.013 & 0.006 & 0.006 & 0.001 & 0.001 \\\hline
    $I_{yy}$    & 0.003 & 0.011 & 0.014 & 0.006 & 0.006 & 0.001 & 0.001   \\\hline
    $I_{zz}$    & 0.003 & 0.005 & 0.004 & 0.002 & 0.002 & 0.001 & 0.001   \\\hline
    $\ell$     & 0.140 & 0.13 & 0.18 & 0.12 & 0.18 & 0 & 0.1      \\\hline
      $m$       &  1  & 4 & 4 & 2.8 & 2.8 & 1.8 & 1.8            \\\hline
\end{tabular}
\end{center}
\end{table}

\section{SIMULATIONS}

A 7-DOF robot was, at first, modeled using a CAD software with the same parameters defined in Table~\ref{tab:1}. The robot is as shown in Fig.\,\ref{fig:2}. This model was then imported to a real-time robotic simulation software named \textit{RoboticsLab v1.6.0}. This real-time simulation software can imitate the experimental conditions coupled with a visual feel of how the robot behaves, which is very important, especially in our cases of position regulation simulations. In this simulation, based on the actual structure of the robot, the centers of gravity of the links are calculated. The friction and gravity effects are included in the simulator environment. The coulomb and viscous friction coefficient values were taken as \textit{unity} for each of the joints. Each joint actuator is modeled as a direct drive gearless light-weight motor with negligible motor inertia while joint sensors are attached for position and velocity measurements respectively. The initial conditions of the robot joints are given in (\ref{eq:20}).
\begin{equation}
    \q_{0} = ( 0^{\circ}, ~90^{\circ}, ~90^{\circ}, ~-90^{\circ} , ~90^{\circ} , ~0^{\circ}, ~90^{\circ}),  ~\qdot_{0} = \mathbf{0}.                  \label{eq:20}
\end{equation}
The simulation in \textit{RoboticsLab} is updated with $1$kHz frequency.

Various simulations are conducted to verify the effectiveness of the proposed method and the results are discussed in detail in the next section. Two sets of simulations are conducted. The first set assumes the nominal model of the robot manipulator as a simple mass-damper model in the form of
${R_N}(s) = {s\left( {\bm{{M_s}^{'}}s + \bm{{B_s}^{'}}} \right)}$ and results are plotted to see the performance of this approach. The second set of simulations is implemented using a detailed non-linear dynamic model as the nominal model of the robot manipulator in the task-space disturbance observer. In this case, ${R_N}(s) = {s\left( {\bm{\Lambda}(\bm{q})s + \bm{\Gamma}(\bm{q},\bm{\dot q})} \right)}$. A comparison between the two results is given in the next section which proves the effectiveness of the proposed scheme.
\section{RESULTS AND DISCUSSION}\label{sec:result}

Simulations are done using the real-time robotic simulation software
\textit{RoboticsLab v1.6.0} using the 7-DOF redundant robot arm. The first set of simulations uses the simple mass-damper nominal model whose parameter values in the three axial directions are tuned appropriately and are given by $2.5$ kg and $1$ kg/s respectively in the X-direction, $2.2$ kg and $0.00001$ kg/s respectively in the Y and Z directions. The cut-off frequency of the Q-filter in the disturbance observer is taken as $20$ Hz.

A Proportional-Derivative (PD) feedback controller, as shown in Fig.\,\ref{fig:0}, is then applied to the end-effector in the cartesian coordinates of the task-space such that the control law is given by $\bm{u} = (K + Bs)\bm{x} + {\overset{\lower0.5em\hbox{$\smash{\scriptscriptstyle\frown}$}}{\bm{f}}_{\bm{d}}}$. The resultant joint torques from the task-space PD-control input are given by $\bm{J}^{T}\bm{u}$. Theoretically, if the values of $K$ and $B$ are tuned properly, then external torques due to unwanted interactions with obstacles, should not affect the robot end-effector position regulation as per (\ref{eq:16}). For this set of simulations, $K=4$ and $B=0.001$ in all the three axial directions $X$, $Y$, and $Z$.

\begin{figure}
    \centering%
    \includegraphics[width=0.8\columnwidth,keepaspectratio,clip]{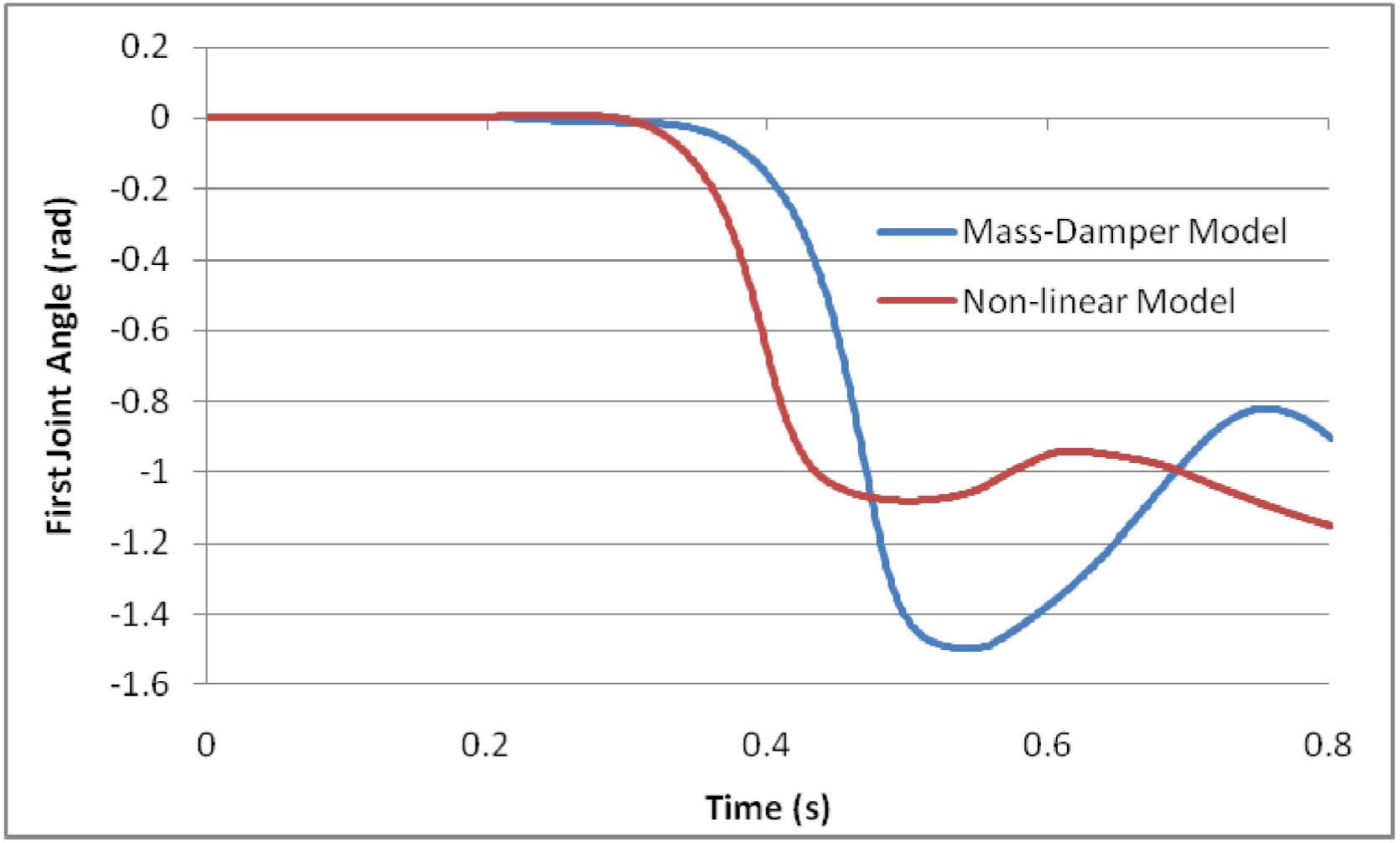}
    \caption{First joint motion using a non-linear and a mass-damper nominal model.}        \label{fig:1}
\end{figure}

\begin{figure}
    \centering%
    \includegraphics[width=0.8\columnwidth,keepaspectratio,clip]{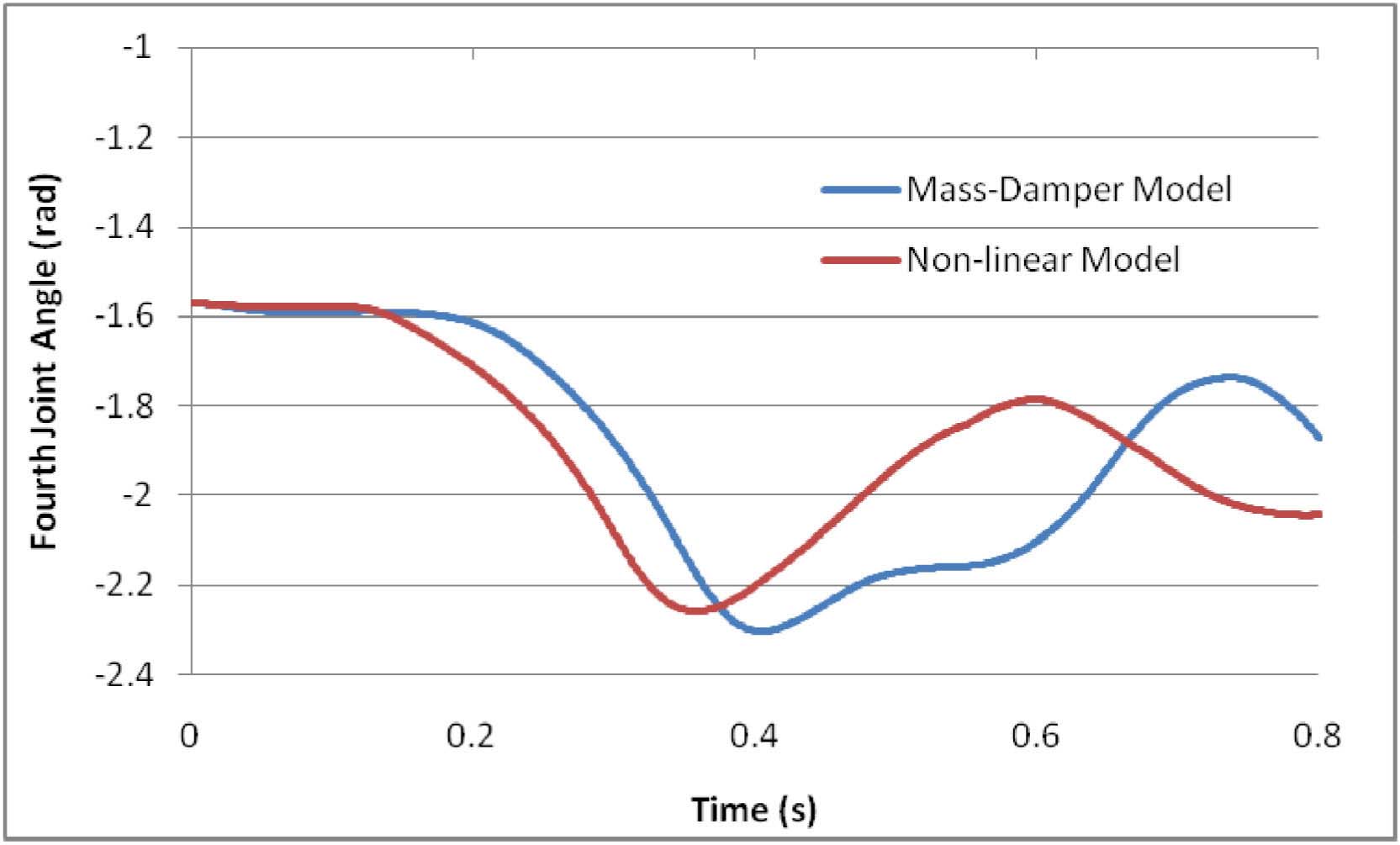}
    \caption{Fourth joint motion using a non-linear and a mass-damper nominal model.}        \label{fig:4}
\end{figure}

\begin{figure}
    \centering%
    \includegraphics[width=0.8\columnwidth,keepaspectratio,clip]{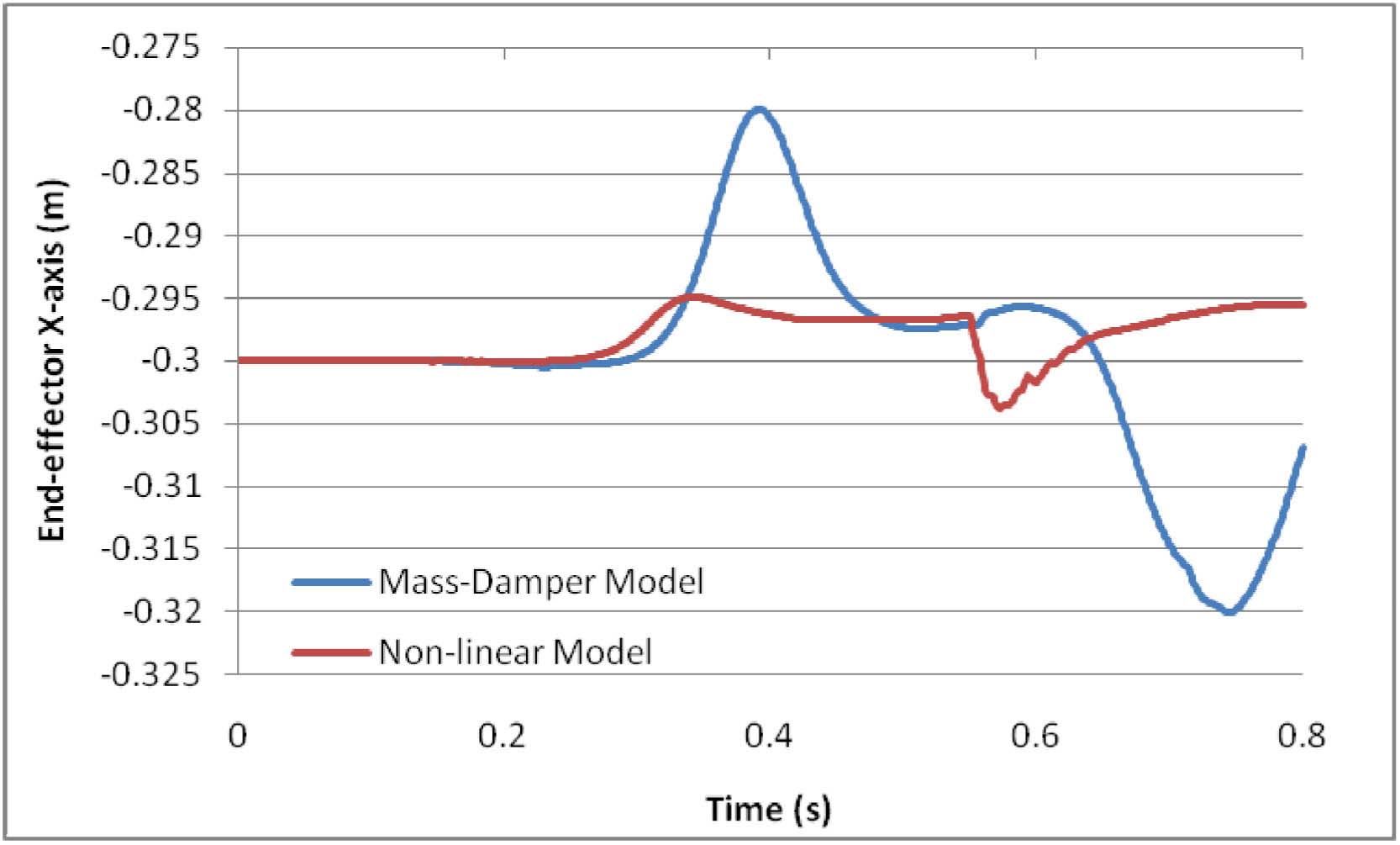}
    \caption{End-effector motion in X-axis using a non-linear and a mass-damper nominal model.}        \label{fig:8}
\end{figure}

\begin{figure}
    \centering%
    \includegraphics[width=0.8\columnwidth,keepaspectratio,clip]{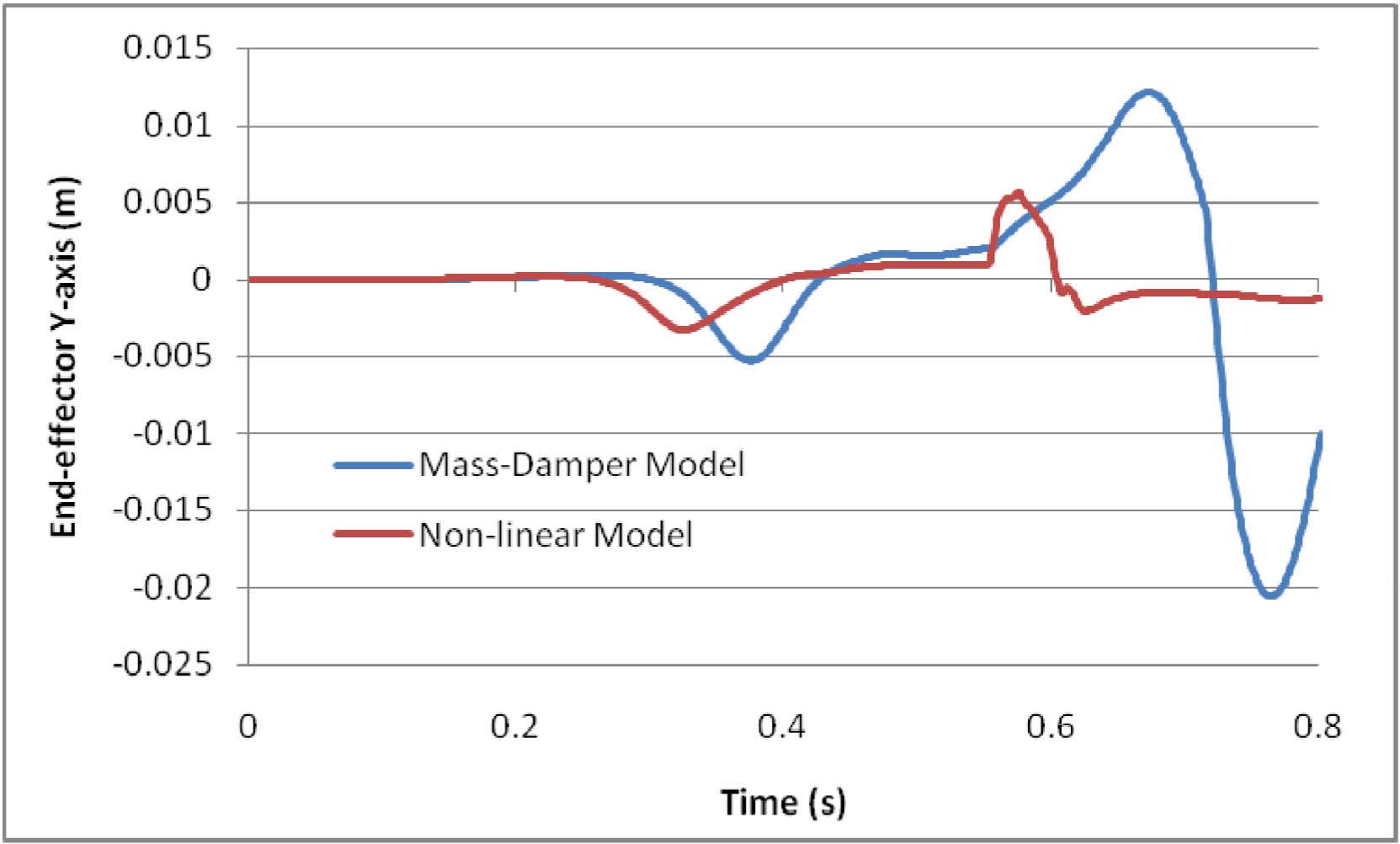}
    \caption{End-effector motion in Y-axis using a non-linear and a mass-damper nominal model.}        \label{fig:9}
\end{figure}

\begin{figure}
    \centering%
    \includegraphics[width=0.8\columnwidth,keepaspectratio,clip]{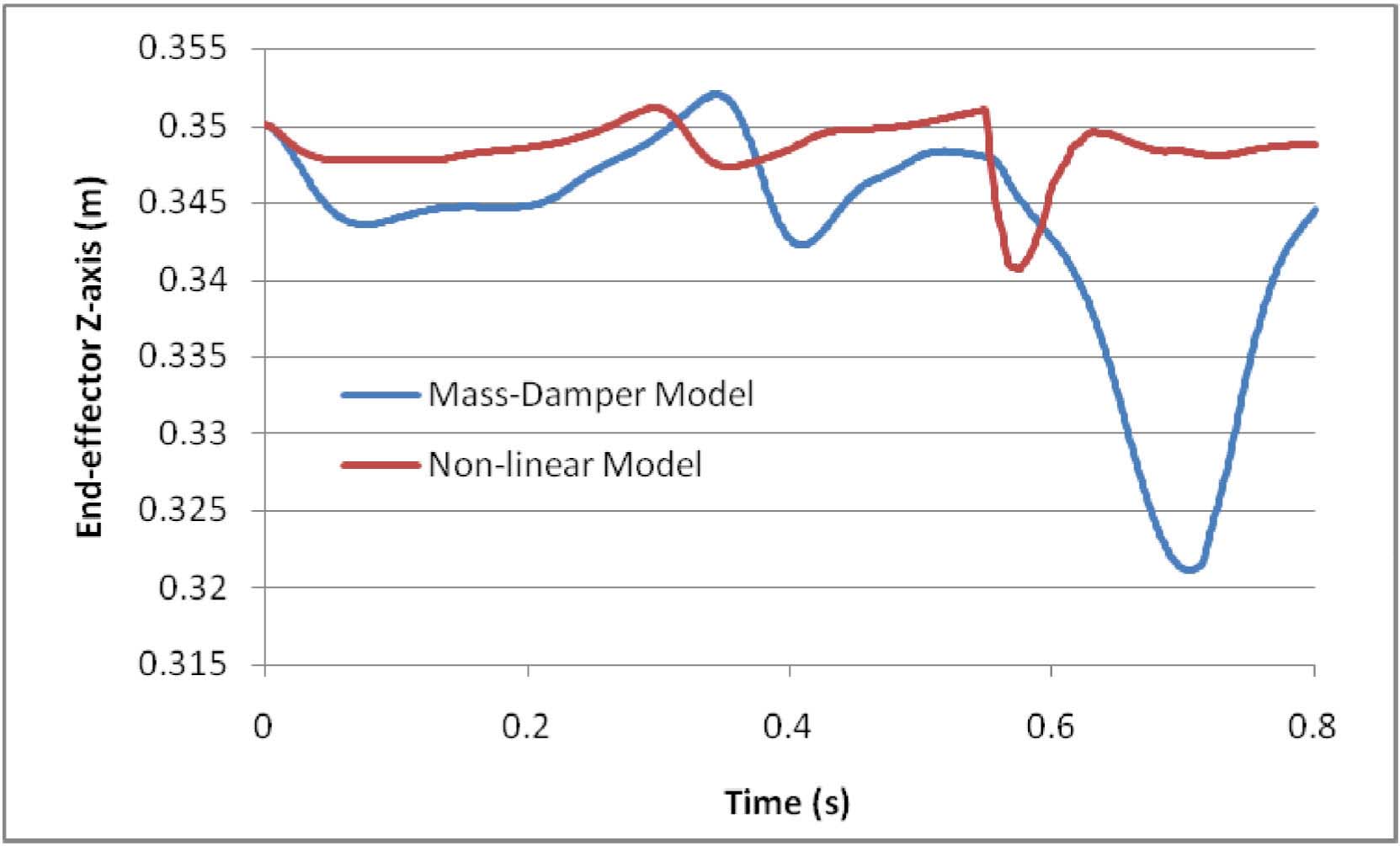}
    \caption{End-effector motion in Z-axis using a non-linear and a mass-damper nominal model.}        \label{fig:10}
\end{figure}

When the simulations are run, the robot manipulator is initially at rest in the absence of any torque/force input as the observer cancels the inertial, coriolis, and gravity effects. Perturbations are then deliberately introduced in the simulation, which is run for around $0.8$s, by applying external forces to different links at different times. Results have been plotted in Figs.\,\ref{fig:1}, \ref{fig:4}, \ref{fig:8}, \ref{fig:9}, and \ref{fig:10}. The choice of the first and fourth joints, to show the motion-profile, is arbitrary and any other choice would show similar results in all circumstances. The standard movement deviation of the robot end-effector for this case is noted to be $0.006$m in the X-direction, $0.0054$m in the Y-direction, and $0.0059$m in the Z-direction. These results show that it is possible to isolate these unwanted perturbations and maintain the end-effector position of any redundant manipulator using this approach. However, position regulation is not so accurate as the motion of the end-effector is not negligible using this mass-damper lumped model.

Simulations, similar to the first set, were then conducted to analyze the effect of choosing a detailed non-linear dynamic nominal model of the robot manipulator in the output of the task-space disturbance observer for position regulation. For this set, the cut-off frequency of the Q-filter, $K$, and $B$ values are same as those of the earlier set.

The computation of the task-space inertia and coriolis matrices are done in accordance with (\ref{eq:7}) which requires the calculation of the derivative of the Jacobian and also the joint-space inertia and coriolis matrices $\bm{M}(\bm{q})$ and $\bm{C}(\bm{q},\bm{\dot q})$ respectively. The computation of these matrices is not possible with the standard Newton-Euler algorithm and hence, a modified Newton-Euler algorithm was used \,\cite{13}. The results are verified by using lagrangian formulation and Christoffel symbols of the second kind\,\cite{9}. The Jacobian derivative was calculated according to \,\cite{14}. The results of this second set of simulations are also plotted in Figs.\,\ref{fig:1}, \ref{fig:4}, \ref{fig:8}, \ref{fig:9}, and \ref{fig:10} for comparison purposes, which show the effectiveness of this newly developed scheme for position regulation.

\begin{figure}
    \centering%
    \includegraphics[width=0.8\columnwidth,keepaspectratio,clip]{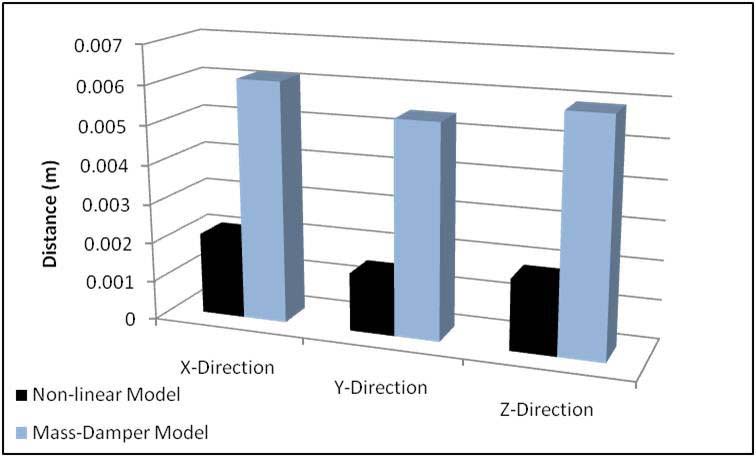}
    \caption{Movement deviation of the robot end-effector for a non-linear and a mass-damper model.}        \label{fig:11}
\end{figure}

Figures \ref{fig:1} and \ref{fig:4} show the motion-profile of the first and the fourth joints of the redundant robot arm respectively for both the lumped model as well as the detailed non-linear model. It can be easily noticed that the motion profile of the first and fourth joints of the redundant robot arm for the detailed non-linear nominal model, as a result of the interactions with obstacles, is quite comparable to that of the simple mass-damper model. Now, if we see Figs.\,\ref{fig:8},  \ref{fig:9}, and \ref{fig:10}, we can note that the end-effector motion, when the detailed non-linear model is used, is negligible  compared to when the lumped model is used. Hence, it should be quite effective for position regulation purposes even if perturbations and non-linear uncertainties are present. The standard movement deviation of the robot end-effector for the case of using the non-linear model is noted to be $0.002$m in the X-direction, $0.0015$m in the Y-direction, and $0.0018$m in the Z-direction. For comparison purposes, it should be noted that the movement of the robot end-effector has reduced by $66.02\%$ in the X-direction, $72.66\%$ in the Y-direction, and $69.40\%$ in the Z-direction due to the use of the non-linear nominal model. The quantitative comparison is shown in Fig.\, \ref{fig:11}.

The simulation video which shows the overall performance of the 7-DOF Robot Arm with the Task-Space Disturbance Observer in presence of external force distrubances is uploaded in \url{https://docs.google.com/open?id=0B5IkR-5za8vGZEtxdFFyUXJTVFU}.

\section{APPLICATION IN HUMAN-LIKE REACHING MOVEMENTS}

\begin{figure}
    \begin{center}
        \includegraphics[width=0.9\columnwidth,keepaspectratio,clip]{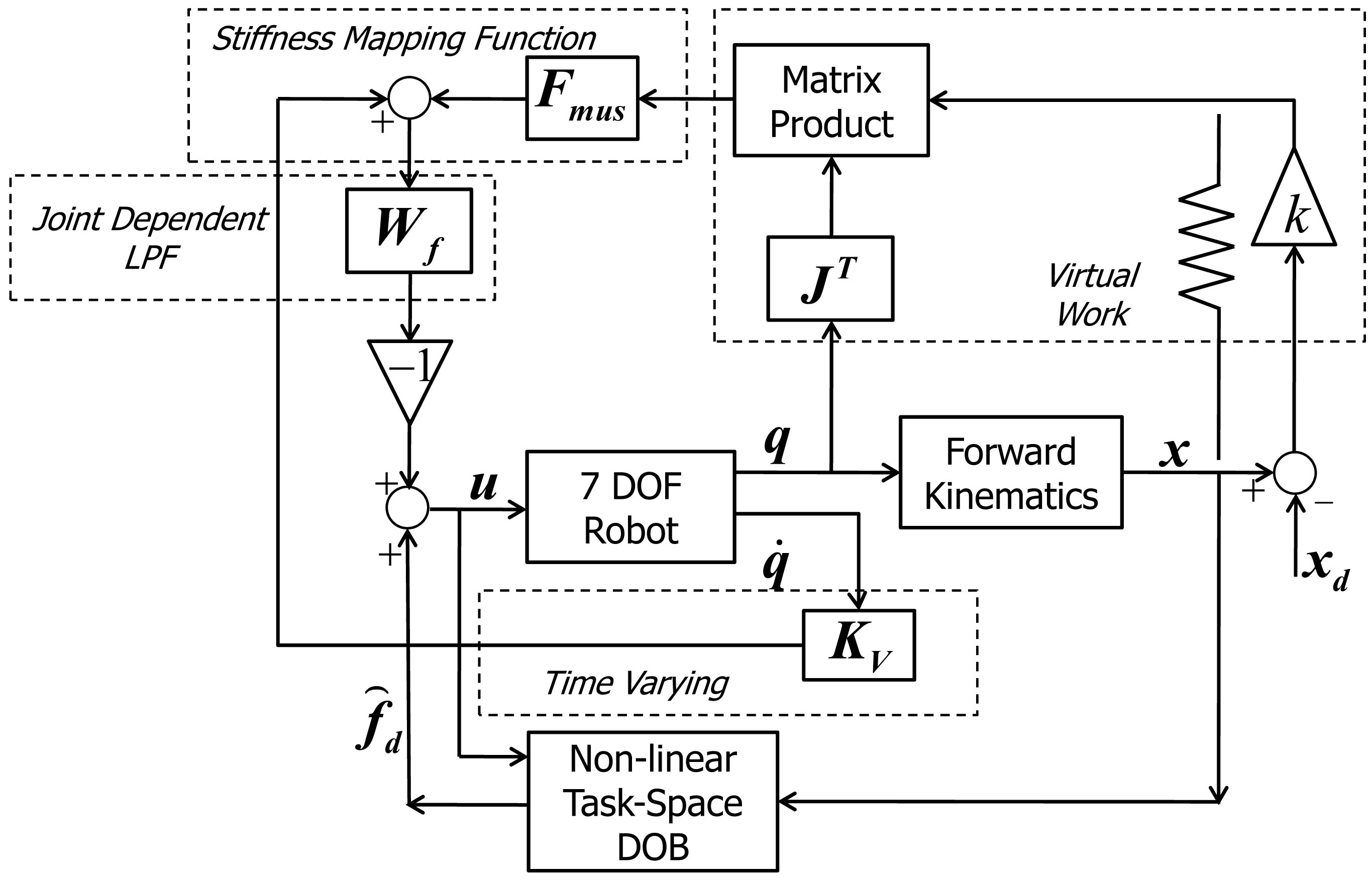}
    \end{center}
    \caption{Block diagram representation of the control law exhibiting human-like characteristics with non-linear task-space disturbance observer.}  \label{fig:control}
\end{figure}

Previously, we had developed a new control law which can exhibit human motion characteristics for reaching movements\,\cite{15}. This control law is modified for the 7-DOF robotic arm system with the inclusion of the non-linear task-space disturbance observer as given below in (\ref{eq:21}).
\begin{equation}
    \u =  - \W_{\bm{f}}\Big[ \K_{\bm{V}}\qdot + k\F_{\bm{mus}}\,\J^{T}(\q)\Delta\x \Big] + \bm{{\overset{\lower0.5em\hbox{$\smash{\scriptscriptstyle\frown}$}}{\f}_{d}}},
                                    \label{eq:21}
\end{equation}
where joint damping matrix $\K_{\bm{V}}$ is given by
\begin{equation}
    \K_{\bm{V}} = {\rm diag}\Big[ C_{1}^{*}(t)~~ C_{2}^{*}(t)~~ ...~~   C_{7}^{*}(t)\Big]   \label{eq:22}
\end{equation}
and
\begin{equation}
    C_{i}^{*}(t) = C_{i}\sin\left( \pi\left( \norm{\Delta\x_{0}}
        - \norm{\Delta\x} \right)/2\norm{\Delta\x_{0}} \right). \label{eq:23}
\end{equation}
$k$ is the virtual spring and $\F_{\bm{mus}}$ is a bijective joint muscle mapping function given by
\begin{equation}
    \F_{\bm{mus}} = {\rm diag}\Big[ f_{1}^{*}(t)~~ f_{2}^{*}(t)~~ ...~~ f_{7}^{*}(t) \Big]   \label{eq:24}
\end{equation}
and
\begin{equation}
    f_{i}^{*}(t) = f_{i}\cos\left( \pi\left( \norm{\Delta\x_{0}}
        - \norm{\Delta\x} \right)/2\norm{\Delta\x_{0}} \right),  \label{eq:25}
\end{equation}
where $f_{i}$ denotes the joint muscle stiffness coefficient and
\begin{equation}
\W_{\bm{f}} = \text{diag}\left[
        \frac{1}{\tau_{i}s + 1} \right],~~\text{for}~i = 1, \ldots, 7\,.
        \label{eq:lpf}
\end{equation}
As given in the modified control law shown in (\ref{eq:21}), the control input to the joint actuators is a function of a damping shaping term given by $\K_{\bm{V}}$ and a bijective joint muscle mapping function given by $\F_{\bm{mus}}$.The damping shaping term $\K_{\bm{V}}$ is assumed to be a diagonal matrix whose elements are time-varying functions. The $\F_{\bm{mus}}$ function on the other hand, is modeled as a time-varying joint dependent bijective muscle mapping function. $\norm{\Delta\x_{0}}$ is the norm of the vector of initial cartesian position error while $\norm{\Delta\x}$ is the norm of the vector of current cartesian position error. This control action is then processed by a low-pass filter matrix, as shown in
(\ref{eq:lpf}), because muscles exhibit inherent low-pass filter characteristics \cite{15}. For a detailed explanation of the control algorithm, please refer \cite{15}. The newly developed non-linear task-space disturbance observer is then applied to this control law and the whole scheme is represented as given in Fig.\, \ref{fig:control}.

\begin{figure}
    \centering%
    \includegraphics[width=0.8\columnwidth,keepaspectratio,clip]{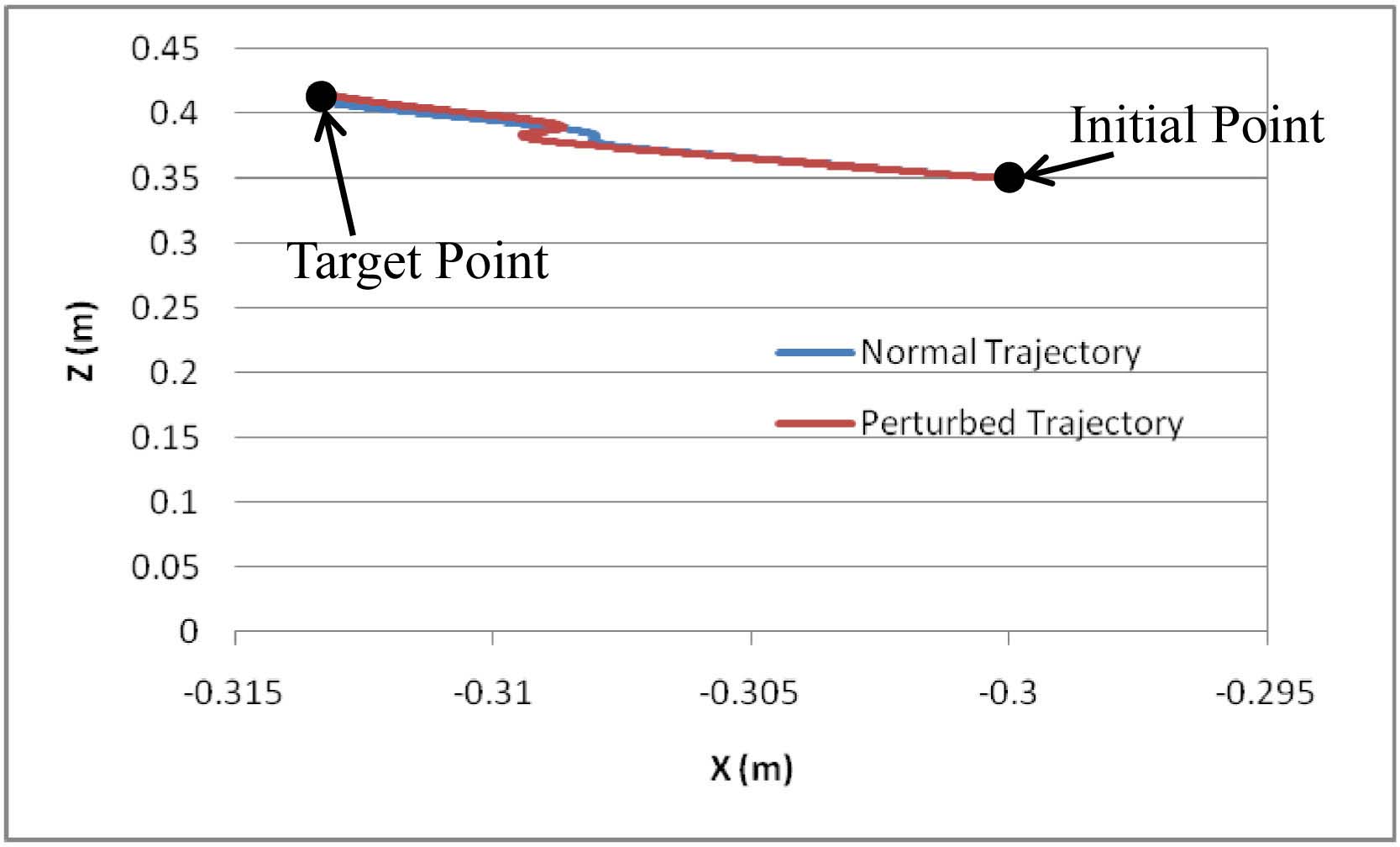}
    \caption{Human-like quasi-straight line trajectory of the robot end-effector for normal and perturbed motion.}        \label{fig:12}
\end{figure}

\begin{figure}
    \centering%
    \includegraphics[width=0.75\columnwidth,keepaspectratio,clip]{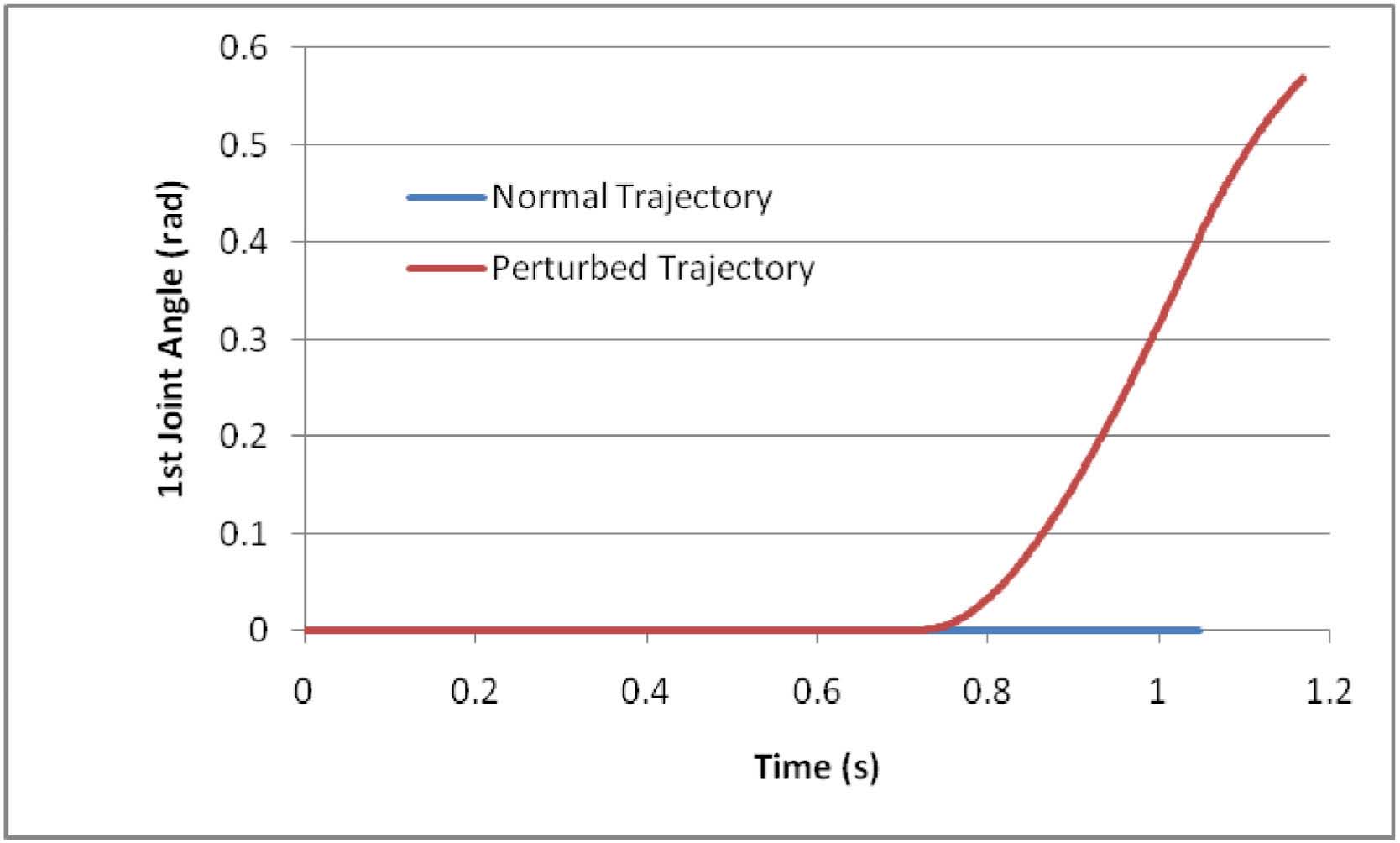}
    \caption{First joint trajectory of the robot arm for normal and perturbed motion.}        \label{fig:13}
\end{figure}

\begin{figure}
    \centering%
    \includegraphics[width=0.75\columnwidth,keepaspectratio,clip]{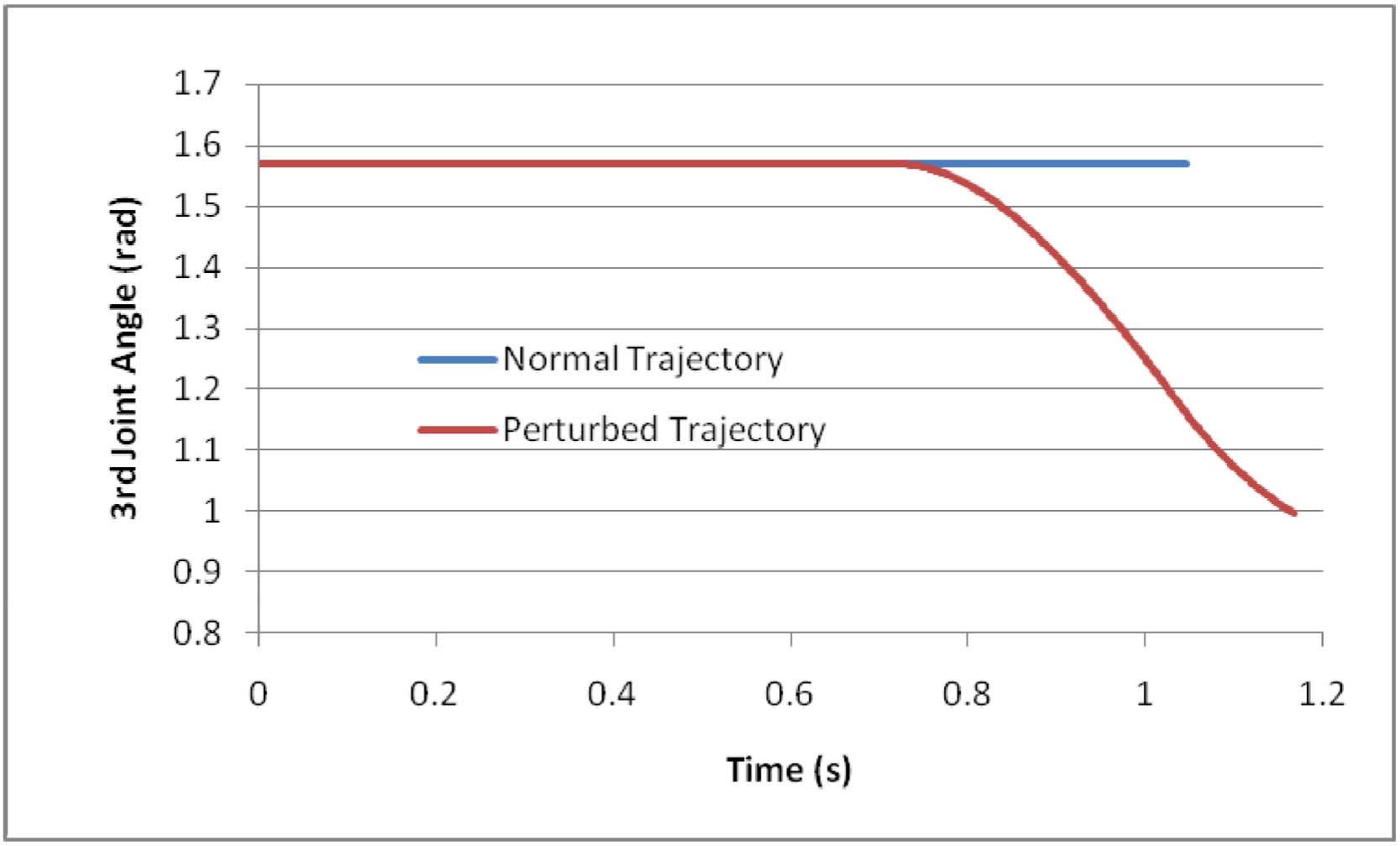}
    \caption{Third joint trajectory of the robot arm for normal and perturbed motion.}        \label{fig:14}
\end{figure}

\begin{figure}
    \centering%
    \includegraphics[width=0.75\columnwidth,keepaspectratio,clip]{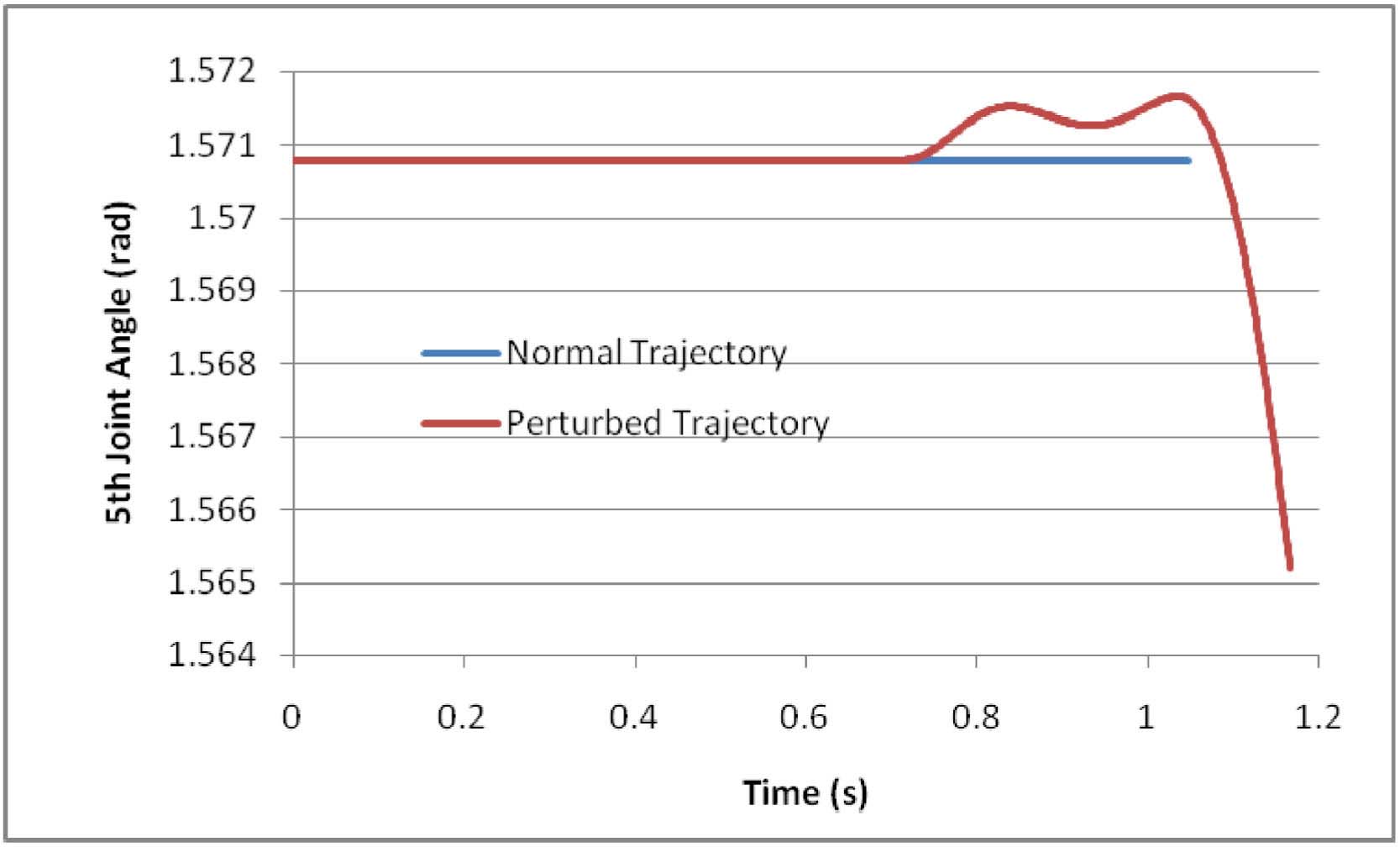}
    \caption{Fifth joint trajectory of the robot arm for normal and perturbed motion.}        \label{fig:15}
\end{figure}

Additional simulations have been conducted using our 7-DOF robot arm to see if the proposed method, using the non-linear task-space disturbance observer, can maintain the human motion characteristics like quasi-straight line trajectory for reaching a target point using (\ref{eq:21}) in spite of perturbations in the environment. The results are shown in Figs.\,\ref{fig:12}, \ref{fig:13}, \ref{fig:14}, and \ref{fig:15}. Firstly, the robot arm is allowed to move towards its target normally without any perturbation. In the second run, perturbations are introduced for the motion towards the same target. The results show that in spite of arbitrary perturbations induced in the environment which greatly affects the 1st, 3rd, and 5th joint trajectory when compared to the normal trajectories as given in Figs.\,\ref{fig:13}, \ref{fig:14}, and \ref{fig:15} respectively, the end-effector trajectory still shows the quasi-straight behavior and deviates negligibly from the normal path as shown in Fig.\,\ref{fig:12}. This further proves the efficacy of our proposed algorithm and shows its high significance in dexterous manipulation using multi-DOF redundant robot arms.

\section{CONCLUSION}

In this report, we have developed a novel method to regulate the position of redundant robot arms against arbitrary perturbations which might be present in 3D environments like unwanted interactions with obstacles in clumsy environments or uncertainties in the robotic system. This method is based on a non-linear task-space disturbance observer which takes into account the detailed non-linear dynamics of a multi-DOF redundant robotic arm system. A modified Newton-Euler algorithm is used to compute this detailed non-linear dynamic model. Simulations have been conducted by using \textit{RoboticsLab} which show the effectiveness of this new method. The results show that this new method can successfully regulate the position of a robot arm in spite of arbitrary perturbations.  These results are then compared with that of a conventional mass-damper based task-space disturbance observer which show the effectiveness of the proposed scheme. It is seen that due to the use of the non-linear model, there is a $66.02\%$ decrease in the robot end-effector movement in X-direction, $72.66\%$ decrease in Y-direction, and $69.40\%$ decrease in the Z-direction in spite of comparable perturbations to the robot arm. This method is then applied to a controller which exhibits human-like motion characteristics for reaching motion. Results show that even if large arbitrary perturbations in the form of unwanted interactions with obstacles are introduced, the end-effector successfully continues to move in the human-like quasi-straight trajectory to reach the target even though the joint trajectories deviated by a considerably large amount. These results are then compared to the unperturbed motion of the same robot arm for the same reaching target which show that in spite of the large differences in the joint trajectories due to these unwanted interactions, the end-effector path remained the same; and this proves the significance of this developed scheme.


\end{document}